# Trial-Based Dominance Enables Non-Parametric Tests to Compare both the Speed and Accuracy of Stochastic Optimizers

Kenneth V. Price, Vacaville, CA USA
Abhishek Kumar, Kyungpook National University, Daegu, South Korea
P. N. Suganthan, KINDI Centre for Computing Research, Qatar University, Qatar and Nanyang Technological University, Singapore

**Abstract**— Non-parametric tests can determine the better of two stochastic optimization algorithms when benchmarking results are ordinal, like the final fitness values of multiple trials. For many benchmarks, however, a trial can also terminate once it reaches a pre-specified target value. When only some trials reach the target value, two variables characterize a trial's outcome: the time it takes to reach the target value (or not) and its final fitness value. This paper describes a simple way to impose linear order on this two-variable trial data set so that traditional non-parametric methods can determine the better algorithm when neither dominates. We illustrate the method with the Mann-Whitney U-test. A simulation demonstrates that "U-scores" are much more effective than dominance when tasked with identifying the better of two algorithms. We test U-scores by having them determine the winners of the CEC 2022 Special Session and Competition on Real-Parameter Numerical Optimization.

## 1. Introduction

Competitions are the venue in which evolutionary algorithms (EAs) vie for supremacy. Ensuring that the fitter EAs survive, however, requires an "objective function" that can identify which of two algorithms is better. In competitions, the objective function typically returns a score derived from the speed and/or accuracy of an algorithm's trials. Because speed and accuracy are conflicting objectives, comparing the performance of stochastic numerical optimization algorithms is a *multi-objective* problem.

Traditionally, there are two complementary approaches to evaluating an algorithm's performance. One is the "fixed target" scenario, which records the number of function evaluations (FEs) that a trial takes to reach a prescribed minimum function error value (EVmin). The other is the "fixed cost" scenario, which records a trial's function error value (EV) after a prescribed maximum number of function evaluations (FEmax). Simply put, data sampled in the fixed-cost scenario allows algorithms to be compared by their EVs, while data sample in the fixed-target scenario allows algorithms to be compared by their FEs. In practice, competitions sample data in a combined scenario by setting both FEmax (to limit computational effort) and EVmin (to avoid rewarding unnecessary accuracy). The problem is how to combine the FEs and EVs recorded in these two different scenarios into a fair metric of performance that can identify the better algorithm regardless of how trials are distributed across the two terminal axes.

For example, Fig. 1 shows six possible ways in which four trials—two from algorithm P and two from algorithm Q—might terminate. Stars denote trial averages. In Figs. 1a–1d, P is the better algorithm *on average* because it is faster (Fig. 1a), more accurate (Fig. 1b) and both faster and more accurate (Fig. 1c and 1d). In each of these cases, we can say that P *dominates* Q because it is better in at least one aspect of performance and not worse in the other. In Figs. 1e and 1f, however, averages cannot determine the better algorithm because neither dominates, i.e. P is faster, but less accurate than Q on average.

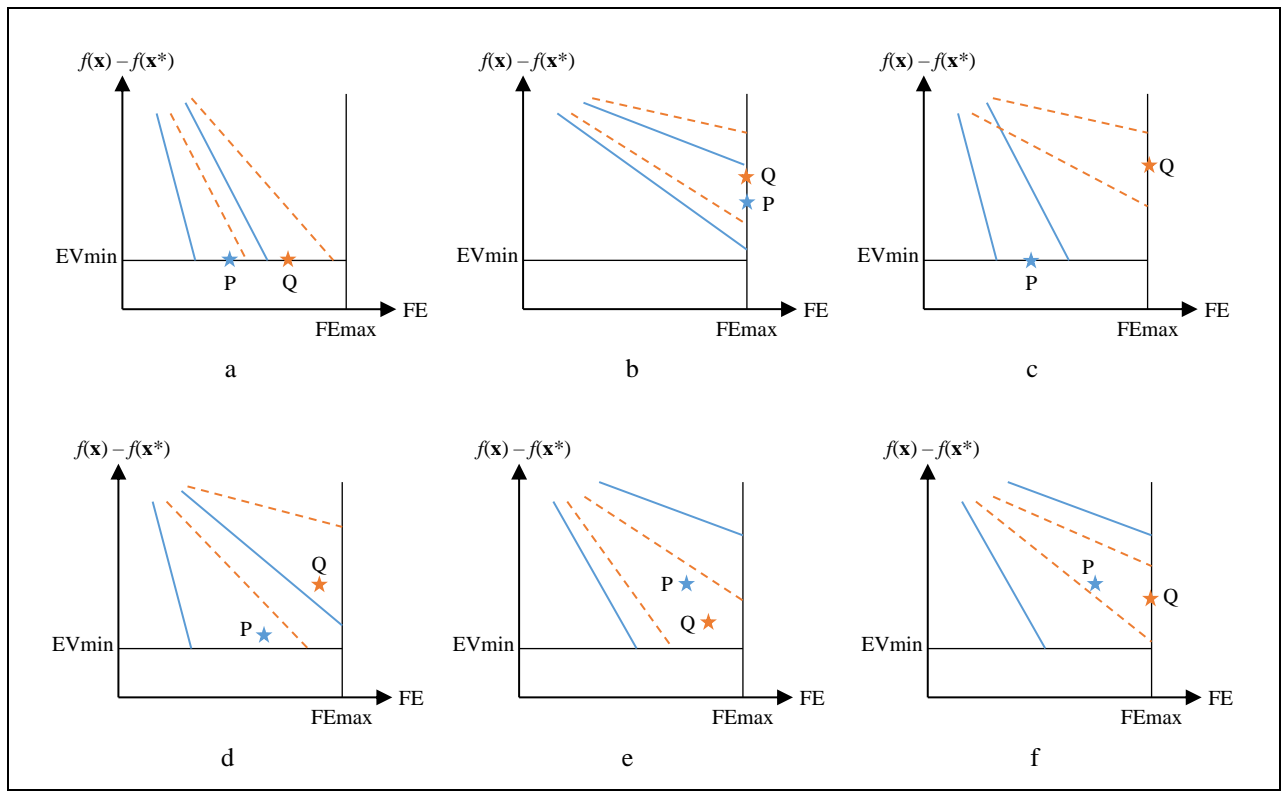

**Fig. 1**. (a) P is faster. (b) P is more accurate. (c) & (d) P is both faster and more accurate. (e) & (f) Neither algorithm dominates. Stars denote average performance.

For a competition between *m* algorithms, dominance that is based on averages can *at most* establish *m* classes, which would allow for algorithms to be unambiguously ranked 1 through *m*. It is, however, far more likely that there will be just a few dominance classes with each containing several algorithms. As a result, there will be many instances in which the better algorithm cannot be determined by averaging trials.

Instead of comparing *average* values, we can refine the dominance classes if we directly *compare all trials*. Specifically, trials can be *unambiguously ranked* from best to worst when they terminate upon reaching either EVmin *or* FEmax. For example, Fig. 2 shows the outcome of four trials, labeled A, B, C and D. Trial A dominates trial B because it was faster to reach EVmin. Trial B dominates trial C because it was both faster and more accurate and trial C dominates trial D because it was more accurate at FEmax. If *smaller is better* (down and to the left in Fig. 2), then we can rank the four trials from best to worst as: A ≻ B ≻ C ≻ D, where "≻" means "dominates".

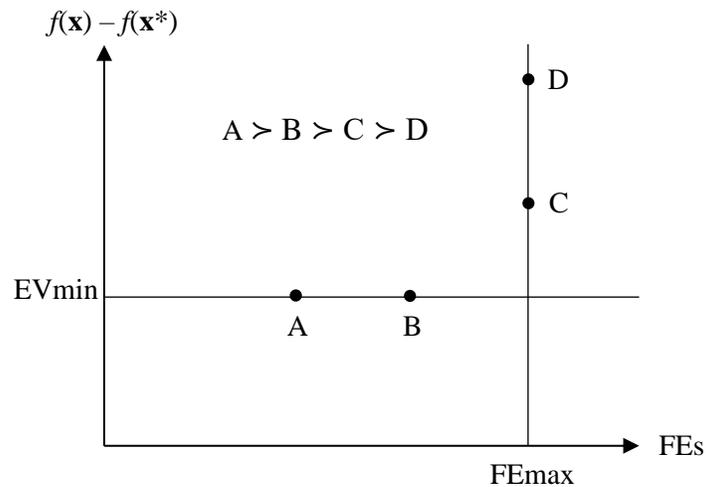

**Fig. 2.** Trials sampled on the terminal axes can be unambiguously ordered from best to worst. Trial A is the best. The symbol "≻" means "dominates".

In a similar way, trials from multiple algorithms can be combined and then unambiguously ranked from best to worse *regardless of whether they terminate at either EVmin or FEmax*. By counting the number of times that an algorithm has the better trial when all of its trials are compared to all trials from all other algorithms, we can fairly characterize an algorithm's comparative performance with a single numerical score without making any assumptions about the relative importance of speed *vs*. accuracy beyond those that are implicit in the choice of terminating criteria. Before detailing how the method works, we offer some perspective by briefly reviewing previous efforts to score the performance of evolutionary optimization algorithms.

## 2. Related Research

Our proposed scoring method cannot evaluate an individual algorithm's performance, but its ability to order two or more algorithms makes it a good choice for competitions and comparative studies where rank is paramount. Consequently, this section focuses on measures of performance that have scored numerical optimization competitions. We do not consider metrics for neither multi-objective nor constrained optimization because, in addition to speed, they require more than one performance indicators —like convergence to/spread over the Pareto front and objective values/constraint violations—respectively, to determine the better algorithm. Additionally, we limit our focus to how competitions have scored two or more algorithms on a *single function*, i.e. we do not extensively consider aggregate scoring over all benchmark functions. Moreover, we disregard *subjective* scoring, like the scheme that was chosen for the Congress on Evolutionary Computation (CEC) 2008 Special Session and Competition on Large Scale Global Optimization (LSGO) [1], which asked participants to rank all algorithms (except their own) based on their expert opinions of the results. The following three sections briefly recount how competitions have scored algorithms based on data recorded in the fixed-target, fixed-cost and combined fixed-target/fixed-cost scenarios.

## 2.1 Fixed-Target: Rewarding Speed

One of the earliest competitive measures of stochastic algorithmic performance dates back to the First International Contest on Evolutionary Optimization (1st ICEO), which was held during CEC 1996 [2]. The 1st ICEO's primary measure of performance was the *expected number of function evaluations per success* (ENES), where a "success" is defined as a trial that reaches a preset function value, aka the "Value-to-Reach" (VTR), or equivalently, a minimum error value EVmin. The ENES has also been the primary performance measure for the Comparing Continuous Optimizers (COCO) benchmarking platform [3] where it was renamed the *expected run-time* (ERT). The Black Box Optimization Benchmark (BBOB) competitions [4] have relied on the COCO platform since 2009.

### 2.1.1 Expected Run-Time

The ERT is defined as the *total* number of function evaluations taken by $n$ trials divided by the number of successful trials $n_s$. If we let $p_s = n_s/n$ be the estimated probability of success, then the ERT can be expressed as a weighted combination of the *average* number of function evaluations for successful $\langle FE^s \rangle$ and unsuccessful $\langle FE^{us} \rangle$ trials, all divided by $p_s$ (Eq. 1).

$$\text{ERT} = \frac{p_s \langle FE^s \rangle + (1 - p_s) \langle FE^{us} \rangle}{p_s}, \quad p_s \neq 0 \tag{1}$$

The main problems with the ERT stem from the way that it handles unsuccessful trials. When all trials fail to reach the VTR, the ERT is undefined. When only some trials are unsuccessful, the ERT not only fails to account for their accuracy, but also becomes strongly dependent on the choice for $FE^{us}$ (and restarts). Both the 1st ICEO and BBOB competitions consider $FE^{us}$ to be an aspect of algorithm design, although both competitions also set FEmax. (FEmax = 1.0e7 for BBOB/COCO, but the 1st ICEO competition left the choice for FEmax to each entrant).

*2.1.2 Empirical Cumulative Distribution Function*

To mitigate the risk of failure and the problems that it causes for the ERT, the BBOB competition samples the number of function evaluations at a series of EV cut-points, the first of which is chosen so that success is virtually assured. The resulting data is then compiled into an *empirical cumulative* (run-time) *distribution function* (ECDF). Specifically, the BBOB competition records the ERT needed to reach 51 different function *error targets*, ranging from 1.0e+2 to 1.0e−8 [5]. On a plot of the ECDF for a given *function*, the abscissa records the ERT, while the ordinate plots the fraction of *targets* reached. (Individual ECDF's for each function in the test bed are then aggregated into a graph that shows the fraction of *function-target* combinations reached as a function of the ERT).

Although the ECDF inherits the ERT's aforementioned deficiencies, it nevertheless provides a more comprehensive view of algorithmic performance. The BBOB competition does not, however, specifically rank algorithms, although it has been proposed that they can be ranked by the *area* under their ECDF curves [6].

*2.2.3 Success Performance*

One way to eliminate the dependence of the ERT on $FE^{us}$ is to base performance on just an algorithm's successful trials. A variant of the ERT, the success performance (SP) [7] is the *average number of function evaluations taken by successful trials* divided by the estimated *probability of success* (Eq. 2). The ERT reduces to the SP when $\langle FE^{us} \rangle = \langle FE^{s} \rangle$.

$$SP = \frac{\langle FE^{s} \rangle}{p_s} \qquad (2)$$

Like the ERT, the SP is undefined when there are no successes and it also fails to account for the error of failed trials. Moreover, the SP requires multiple successful trials to have any statistical significance. Similar to BBOB and the ERT, the CEC 2005 Special Session and Competition on Real-Parameter Optimization (RPO) [8] compiled an empirical cumulative distribution function of the SP for all algorithms to more completely assess their performance. (Section 2.3 contains additional details about this competition's evaluation criteria).

**2.2 Fixed-Cost: Rewarding Accuracy**

Before being divided by $p_s$, the ERT is the average number of FEs taken by all trials, while the SP is the average number of FEs taken by just the successful trials. Neither measure, however, accounts for the *error* of *failed* trials. By contrast, the scores described in this section record the error of *all* trials, but they do not distinguish successful trials by their speed.

In most competitions, participants recorded five *error data* (ED) points: the *best*, *median*, *worst* and *mean* function values, along with the *standard deviation* of multiple trials at one or more FE cut-points for each function, but not all competitions incorporated all five ED points into an algorithm's function score. In most cases, *function* scores were a *sum* of *category* scores, where a category is an (ED, FE) combination. Basically, competitions computed category scores as:

- a rank,
- a function of rank,
- an error value, or
- a hybrid combination of both ranks and error values.

We first look at methods whose category scores were ranks.

*2.2.1 Rank*

The CEC 2011 Competition on Testing Evolutionary Algorithms on Real-World Problems [9] ranked algorithms by both their *mean* and *best* values at FEmax (two categories), then summed the algorithm's two category ranks to determine its function score.

The scheme chosen for the CEC 2013 Competition on RPO [10] most closely resembles our own. Like the Wilcoxon Rank-Sum test [11], it combined all trials from multiple algorithms and ranked them based on their final EVs. An algorithm's function score was just its Wilcoxon R-statistic, i.e. the sum of the ranks of its trials. (By contrast, our method ranks trials based on both their EVs *and* FEs before summing ranks and subtracting a constant so that the statistic represents the number of "wins" that an algorithm accumulates).

*2.2.2 Function of Rank: Formula 1*

The CEC 2010 and 2012 LSGO competitions [12], [13] sampled all five ED points at three FE cut-points for a total of 15 categories. After ranking each algorithm in each category based on its ED value (smaller is better in every case), these LSGO competitions assigned a *Formula 1* (F1) score to each rank. Table 1 shows that F1 scoring disproportionately rewards higher ranks. In these competitions, an algorithm's function score was the sum of its F1 scores over all 15 categories.

Table 1. Formula 1 scores as a function of rank, $R$.

| $R$ | 1 | 2 | 3 | 4 | 5 | 6 | 7 | 8 | 9 | 10 | >10 |
|---|---|---|---|---|---|---|---|---|---|---|---|
| F1($R$) | 25 | 18 | 15 | 12 | 10 | 8 | 6 | 4 | 2 | 1 | 0 |

The CEC 2013 and 2015 LSGO competitions [14], [15] only ranked algorithms based on their *median* error value at each of three FE cut-points for a total of three categories. Like their earlier incarnations, these LSGO competitions also transformed ranks into F1 scores before summing them for a function score.

The CEC 2014 Competition on Computationally Expensive Optimization [16] ranked algorithms based on both their *mean* and *median* error values at FEmax (two categories) and then awarded 9, 6 and 3 points to ranks 1, 2 and 3, respectively, with no points being awarded to lower ranks.

*2.2.3 Error Value*

The CEC 2015 Competition on Computationally Expensive Optimization [17] sampled both the *mean* and *median* error at two FE cut-points (four categories). One half of an algorithm's score was the average of the *mean* EVs sampled at the two FE cut-points. The other half was based on the average of the *median* EVs sampled at the same two FE cut-points.

The CEC 2019 100-Digit Challenge [18] ranked algorithms by the average number of correct digits (up to 10) that they found in the best 25 out of 50 trials. The choice for FEmax was left to each competitor.

*2.2.4 Hybrid Scores: Ranks and Error Values*

The score developed for the CEC 2017 Special Session and Competition on Single Objective Bound Constrained Numerical Optimization [19] was a 50/50 combination of two separate measures of function error: the sum of errors (SE) and the sum of ranks (SR). An algorithm's contribution to the SE score for a given function was its final *mean* error value. The contribution to an algorithm's SR score for a given function was its *rank* among other algorithms based on its final *mean* error value.

The CEC Special Sessions and Competition on Single Objective Bound Constrained Numerical Optimization in 2020 and 2021 [20], [21] normalized the error scores explored at CEC 2017 so that each function contributed more equitably to the total SE score (before being multiplied by a dimension-dependent weight).

**2.3 Fixed-Cost and Fixed-Target: Rewarding Success or Accuracy**

The ranking criterion developed for the CEC 2005 Competition on Real Parameter Optimization [6] resembles our scoring scheme in that it also relied on data that were sampled in both the fixed-cost and fixed-target scenarios. For a given function, the better of two algorithms was the one with:

- the *most successes* when at least one algorithm had one or more successful trials.
- the lower *median error* when neither algorithm had a successful trial.

This approach fairly ranks algorithms by their median EVs when neither has a success, but it does not account for the error of failed trials when an algorithm has at least one success. Additionally, this scheme does not reward the faster of two equally successful algorithms. While the SP can distinguish between the faster of two equally successful algorithms, a less successful algorithm might have a better SP, i.e. the most successful algorithm did not always have the best SP. The point here is that one can rank successful algorithms either by their number of successes, or by their SP, but not both without risking conflicts.

See [22] and [23] for additional perspectives on both the CEC and BBOB competitions. Many CEC competition results can be found online at: https://github.com/P-N-Suganthan and also at the Power Engineering Database [24]: https://www.al-roomi.org/benchmarks/cec-database

## 3. U-Scoring

With the exception of the CEC 2013 RPO competition, the Wilcoxon rank-sum R-statistic has not traditionally been a metric for competitions, even though it is an accepted standard for comparing two algorithms when a *single* variable characterizes their performance. Our extension to *two* variables works because the rank-sum method only requires that data be ordinal. As shown in the introduction, it is possible to *unambiguously* determine which is the better of two *trials* regardless of whether they terminate upon reaching EVmin or FEmax. In this way, trials from multiple algorithms can be combined and then ranked from best to worst. Summing the ranks of an algorithm's trials and subtracting a constant gives the Mann-Whitney U-statistic [25], which is the number of wins that the algorithm scores in a comparison with all trials from all other algorithms (on a given function).

### 3.1 Comparing *m* Algorithms on One Function

Suppose that $t_{i,j,k}$ is the $i^{\text{th}}$ trial from the $j^{\text{th}}$ algorithm optimizing the $k^{\text{th}}$ function. To compute the algorithms' U-scores for function $k$, rank all trials $t_{i,j,k}$, $i = 1, 2\ldots, n$, $j = 1, 2\ldots, m$, where $n$ is the number of trials and $m$ is the number of algorithms. Assign the best trial the highest rank ($nm$). Resolve ties by assigning average ranks to identical trials. Once trials have been ranked, compute an algorithm's U-score for function $k$ as the *sum of the ranks of its trials* minus the correction term

$n(n + 1)/2$ (so that algorithms do not count their own trials as wins). Figure 2 provides an example with three algorithms, P, Q and R, each of which ran four trials on one function. Table 1 illustrates how to compute algorithm U-scores from Fig. 2. Although more computationally expensive, it is usually simpler to compute U-scores by comparing all (unranked) trials from all algorithms (on one function) and counting the number of wins that each algorithm receives.

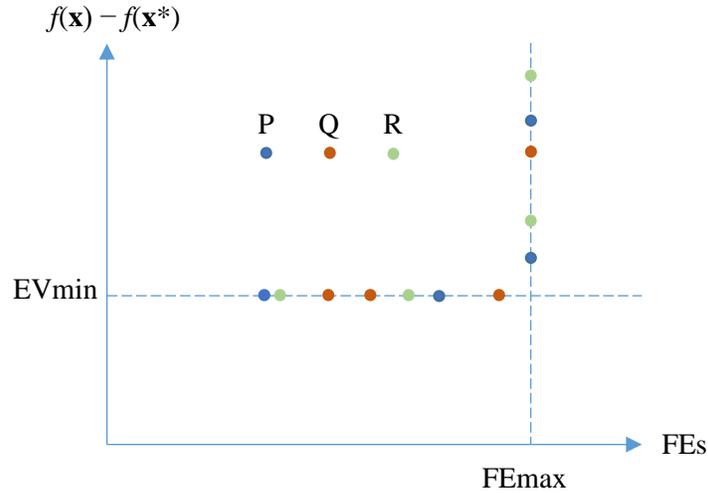

**Fig. 2.** Three algorithms, P, Q and R, run four trials each. Five trials terminate when they reach FEmax, while seven trials terminate when they reach EVmin. All twelve trials can be ordered from best to worst.

**Table 2.** Function scores for algorithms P, Q and R are derived by summing their ranks. Algorithm Q wins with a score of 18. The correction factor for four trials is 4·5/2 = 10. "SR" = sum of ranks.

| Trial: | p | r | q | q | r | p | q | p | r | q | p | r | SR | U-score |
|---|---|---|---|---|---|---|---|---|---|---|---|---|---|---|
| Rank: | 12 | 11 | 10 | 9 | 8 | 7 | 6 | 5 | 4 | 3 | 2 | 1 | 78 | — |
| P | 12 |   |   |   |   | 7 |   | 5 |   |   | 2 |   | 26 | 16 |
| Q |   |   | 10 | 9 |   |   | 6 |   |   | 3 |   |   | 28 | 18 |
| R |   | 11 |   |   | 8 |   |   |   | 4 |   |   | 1 | 24 | 14 |

In competitions, ranking is paramount and the statistical significance of the difference in ranks is a secondary concern. Nevertheless, when only two algorithms compete, U-scores reduce to the traditional Mann-Whitney U-statistic for determining statistical significance. Tables 3–5 decompose Table 2 into three, pair-wise comparisons: P *vs*. Q, P *vs*. R and Q *vs*. R. The difference in performance between two algorithms is statistically significant if the lower of the two U-scores is less than or equal to a critical value from a look-up table that is indexed by both the number of trials and the level of significance. (None of the differences in this example were significant at the 0.05 level in a two-tailed Mann-Whitney U-test with a sample size of 4). Table 6 shows that the sum of an algorithm's pairwise U-scores equals its U-score in Table 2.

Table 3. U-scores for algorithms P vs. Q.

| Trial: | p | q | q | p | q | p | q | p | SR | U-score |
|---|---|---|---|---|---|---|---|---|---|---|
| Rank: | 8 | 7 | 6 | 5 | 4 | 3 | 2 | 1 | 36 | — |
| P | 8 |   |   | 5 |   | 3 |   | 1 | 17 | 7 |
| Q |   | 7 | 6 |   | 4 |   | 2 |   | 19 | 9 |

Table 4. U-scores for algorithms P vs. R.

| Trial: | p | r | r | p | p | r | p | r | SR | U-score |
|---|---|---|---|---|---|---|---|---|---|---|
| Rank: | 8 | 7 | 6 | 5 | 4 | 3 | 2 | 1 | 36 | — |
| P | 8 |   |   | 5 | 4 |   | 2 |   | 19 | 9 |
| R |   | 7 | 6 |   |   | 3 |   | 1 | 17 | 7 |

Table 5. U-scores for algorithms Q vs. R.

| Trial: | r | q | q | r | q | r | q | r | SR | U-score |
|---|---|---|---|---|---|---|---|---|---|---|
| Rank: | 8 | 7 | 6 | 5 | 4 | 3 | 2 | 1 | 36 | — |
| Q |   | 7 | 6 |   | 4 |   | 2 |   | 19 | 9 |
| R | 8 |   |   | 5 |   | 3 |   | 1 | 17 | 7 |

Table 6. Summing the pair-wise U-scores gives the same U-scores found in Table 2.

| vs. | P | Q | R |
|---|---|---|---|
| P | — | 9 | 7 |
| Q | 7 | — | 7 |
| R | 9 | 9 | — |
|   | 16 | 18 | 14 |

## 3.2 Ties

When scoring two or more algorithms on a single function, ties *may* be possible. The simplest case is for two algorithms ($m = 2$) running two trials each ($n = 2$): p q q p (or q p p q), for which the U-score for P is 2 + 0 = 2 and the U-score for Q is 1 + 1 = 2. More generally, ties are possible whenever the number of comparisons $n^2m(m-1)/2$ is evenly divisible by an integer in the range [2, $m$]. For example, if there are two algorithms, then the number of comparisons is just $n^2$, so ties are possible whenever $n^2$ is divisible by 2, i.e. when $n$ is even. It follows that there can be no ties if $m = 2$ and $n$ is odd. (This analysis excludes the possibility that two trials are exactly the same, with each receiving half a point—something that never occurred during many millions of simulations, but did occur in an actual competition. See section 5.3 for an explanation).

The case for $m = 3$ is more complex because in addition to two-way ties, there can also be three-way ties. More particularly, when $m = 3$, two-way ties can occur if $3n^2$ is divisible by 2, i.e. when $n$ is even, so like the case for $m = 2$, there can be no two-way ties if $n$ is odd. By contrast, the less probable three-way ties are always possible because $3n^2$ is always divisible by 3.

As $m$ increases, determining the probability of a tie becomes increasing complex. Nevertheless, a judicial choice for $n$ can eliminate ties in some cases. For example, if $m = 10$, then ties are possible when $45n^2$ is divisible by integers in the range [2, 10]. Since $45 = 3^25$, ties between 3, 5 and 9 algorithms will always be possible regardless of the choice of $n$. This leaves denominators 2, 4, 6, 7, 8 and 10. If $n$ is odd, then its square cannot be evenly divided by 2, 4, 6, 8 or 10, which leaves just 7. Consequently, we can *minimize* the possibility of a tie for ten algorithms by choosing $n$ odd and not divisible by 7.

### 3.3 Ranking Functions by Difficulty

Like gauging algorithmic performance, quantifying a function's difficulty is not straightforward and yet, by *exchanging the roles* of algorithm and function, U-scores can also rank two or more *functions* with the same dataset that determines algorithmic U-scores. For example, we can compare two functions in the same way that we compare two algorithms by comparing all trials from all algorithms that were run on one function with all trials from all algorithms that were run on another *function*. The function that receives the most wins is the *easier* function. Like their algorithmic counterparts, *functional* U-scores (not to be confused with an algorithm's *function* U-score) can be granular enough to give the relative degree of difficulty, i.e. a sense of *how much* more difficult one function was than another for the given suite of algorithms.

## 4. Simulations

When the better algorithm can be decided by a single variable's average, like average speed (Fig. 1a) or average accuracy (Fig. 1b), then U-scores should closely agree. U-scores should also agree about which is the dominant algorithm when two variables characterize an algorithm's trials (Figs. 1c, 1d). Finally, we expect that U-scores can reliably determine the better of two algorithms in a two-variable scenario even when neither algorithm dominates (Figs. 1e, 1f).

To better understand how well U-scores satisfy these demands, we simulated trial data for algorithms P and Q with normal distributions. First, we simulated a one-parameter scenario to show how the better algorithm is determined and to find out how often U-scores disagree with the determination based on trial averages. We subsequently simulated a two-parameter scenario to investigate how well both dominance and U-scores were able to identify the better algorithm.

### 4.1 The Single Parameter Scenario

We simulated trials for algorithms P and Q with values $p_i$ and $q_i$, $i = 1, 2\ldots, n$, that were drawn from different normal distributions (Eq. 3). Although the standard deviation for both distributions was 1.0, their means, $\mu_P$ and $\mu_Q$, were different by a variable separation $s$ such that $\mu_P = -s/2$ and $\mu_Q = s/2$ (Fig. 3).

$$p_i \sim \mathcal{N}\left(\frac{-s}{2}, 1.0\right), \quad q_i \sim \mathcal{N}\left(\frac{s}{2}, 1.0\right), \quad i = 1, 2 \ldots, n. \tag{3}$$

If, like function evaluations and error values, *smaller is better*, then P is better than Q according to their *true means* when $s > 0$ since $\mu_P < \mu_Q$, but if the better algorithm is determined by just the *sample averages* (Eq. 4), then statistical fluctuations occasionally make it possible for Q to be better than P.

$$\langle p \rangle = \frac{1}{n}\sum_{i=1}^{n} p_i, \quad \langle q \rangle = \frac{1}{n}\sum_{i=1}^{n} q_i \tag{4}$$

Consequently, the experiments below assume that the *sample averages* defined in Eq. 4 *define the better algorithm*.

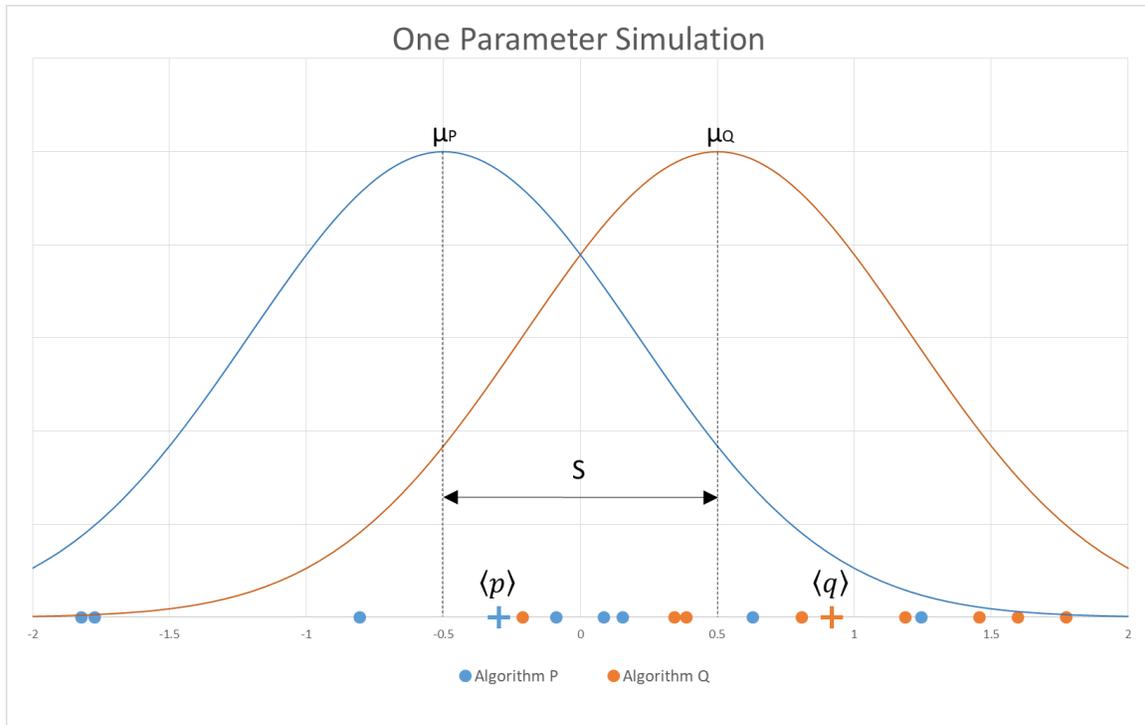

**Fig. 3.** Simulated trials were normally distributed with standard deviation 1.0, but different means separated by *s*. Crosses indicate average performance. Algorithm P (blue) is better than algorithm Q (orange) because $\langle p \rangle < \langle q \rangle$ (smaller is better).

## 4.2 U-scores vs. Trial Averages

This experiment explored how often U-scores and sample averages disagreed about which algorithm was better. We not only counted a "miss" when U-scores disagreed with sample averages, but also when U-scores for P and Q were equal (because there were no cases for which $\langle p \rangle = \langle q \rangle$).

As Fig. 4 shows, U-scores disagreed with 30-trial sample averages less than 8% of the time when $s = 0$. Approximately, 0.6% of those cases were ties. (As per section 3.3, *two* algorithms with an *odd* sample size would have yielded *no* ties). As the separation between the two distributions grows, the probability of a disagreement between sample averages and U-scores exponentially decreases according to a normal distribution. Figure 5 shows how the probability of U-scores disagreeing with sample averages drops as the number of trials in the sample doubles from 15 to 240 trials.

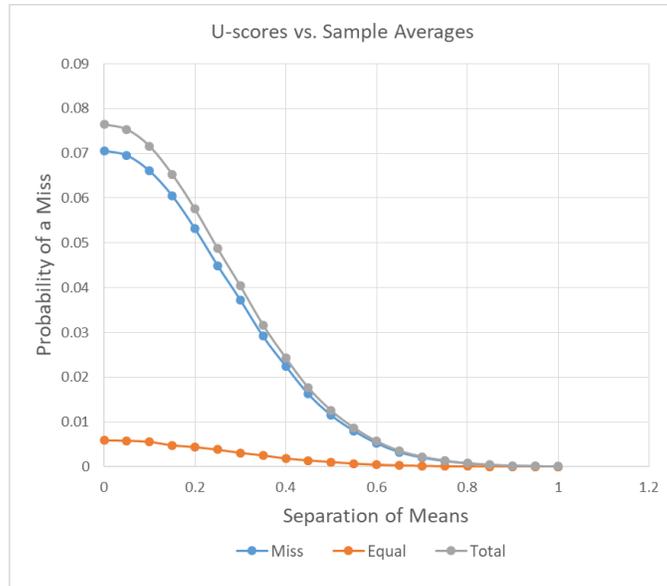

**Fig. 4**. U-scores disagreed with sample averages less than 8% of the time when $\mu_P = \mu_Q$ ($s = 0$), then quickly dropped.

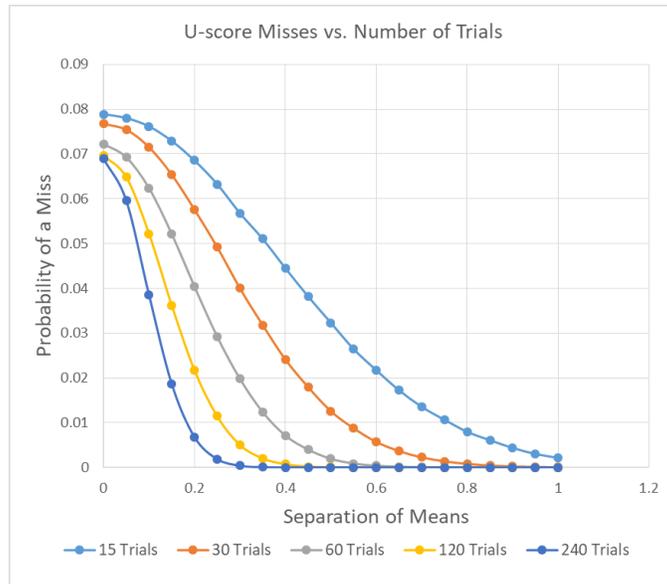

**Fig. 5**. Total U-score misses with respect to sample averages as a function of $s$ and the number of trials.

### 4.3 The Dual Parameter Scenario

Starting with values generated according to Eq. 3, we simulated a two-parameter scenario by dividing the range of values into two domains. Equations 5 and 6 show how values above 0.0 were considered akin to error values $pe_i$ and $qe_i$, while those less than or equal to 0.0 were treated separately as though they were function evaluations $pf_i$ and $qf_i$. In effect, the positive values on the abscissa are rotated counterclockwise by 90 degrees to become positive values on the ordinate (Fig. 6).

$$\begin{aligned} &\text{if } p_i \leq 0 \quad pe_i = 0; \; pf_i = p_i \\ &\text{else} \qquad\quad pe_i = p_i; \; pf_i = 0 \end{aligned} \tag{5}$$

$$\begin{aligned} &\text{if } q_i \leq 0 \quad qe_i = 0; \; qf_i = q_i \\ &\text{else} \qquad\quad qe_i = q_i; \; qf_i = 0 \end{aligned} \tag{6}$$

Now, two averages characterize each algorithm's performance. We denote the *average error values* as $\langle pe \rangle$ and $\langle qe \rangle$ and the *average number of function evaluations* as $\langle pf \rangle$ and $\langle qf \rangle$ for algorithms P (Eq. 7) and Q (Eq. 8), respectively.

$$\langle pe \rangle = \frac{1}{n}\sum_{i=1}^{n} pe_i, \quad \langle pf \rangle = \frac{1}{n}\sum_{i=1}^{n} pf_i \tag{7}$$

$$\langle qe \rangle = \frac{1}{n}\sum_{i=1}^{n} qe_i, \quad \langle qf \rangle = \frac{1}{n}\sum_{i=1}^{n} qf_i \tag{8}$$

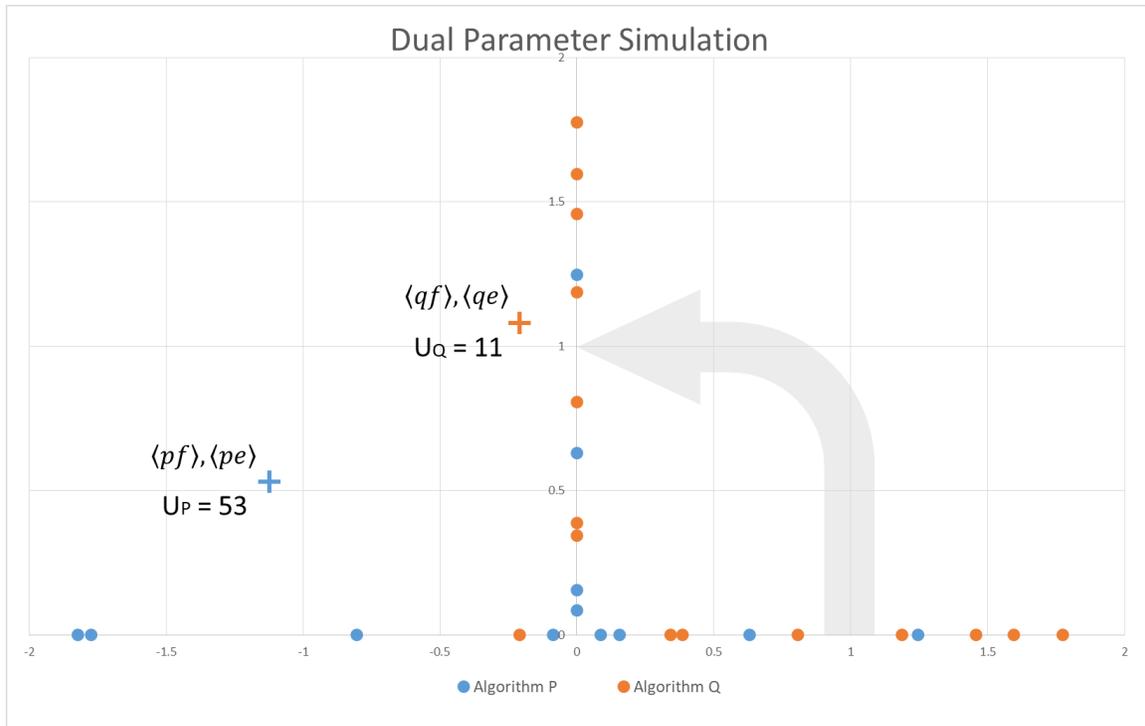

**Fig. 6**. After rotation, simulated "FE" values on the abscissa that are greater than 0.0 become "EV" values on the ordinate. In this example, algorithm P is the better algorithm because it both outscores and dominates algorithm P.

Even though the single axis model is transformed into a dual axis one, P is still the better algorithm as *s* increases simply because it becomes faster at a fixed EVmin, while Q becomes less accurate at a fixed FEmax. In other words, just because P does not dominate based on the dual averages ($\langle pe \rangle, \langle pf \rangle$) and ($\langle qe \rangle, \langle qf \rangle$), doesn't mean that it isn't the better algorithm (in a large sample). Additionally, rotating the positive abscissa to the positive ordinate has no effect on U-scores because it does not change the order of the trials. Even if one of these two ranges were assigned a different scale, U-scores would not change because *order-preserving transforms cannot alter a trial's rank*.

## 4.4 Dominance vs. U-Scores

This experiment computed how often, as a function of *s*, U-scores and dominance misidentified the better algorithm. We counted a "miss" whenever the decision based on either U-scores or dominance disagreed with the one-parameter sample averages, $\langle p \rangle$ and $\langle q \rangle$. It is important to note that if the dual parameter averages show that one algorithm dominates the other, then the one-parameter averages $\langle p \rangle$ and $\langle q \rangle$ will agree that the dominant algorithm is also the better algorithm. Consequently, *the only misidentifications made by the dual average dominance criterion in this experiment are those cases in which neither algorithm dominates*.

Figure 7 shows the results. The probabilities for each value of *s* = 0.0, 0.05, 0.1…, 1.0 are averaged over $10^6$ runs, where each run is a sample of 30 trials per algorithm. The upper and lower curves are the probability that dominance and U-scores, respectively, misidentified the better algorithm. The difference between the two curves represents those cases for which U-scores *correctly identified the better algorithm that dominance was unable to categorize*. Thus, this simulation shows that U-scores are much more effective than average-based dominance at identifying the better algorithm in a two-variable scenario.

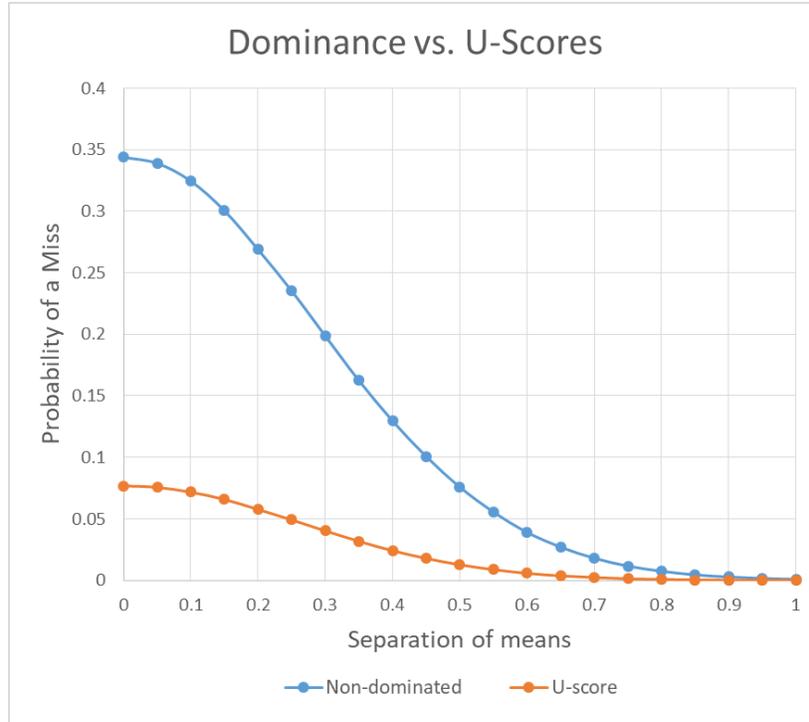

**Fig. 7**. The probability that neither algorithm dominates based on dual averages (blue) and the probability that U-scores fail to identify the better algorithm based on simple averages (orange).

## 5.0 CEC 2022: A Real-World Test

We tested the effectiveness of U-scores in a real-world scenario by having them determine the winners of the 2022 CEC Special Session and Competition on Single Objective Bound Constrained Numerical Optimization [26]. The next section presents that competition's testbed, while the subsequent section lists the algorithms that entered the CEC 2022 competition. A short discussion follows the presentation of the results.

### 5.1 Benchmark Functions

The functions for the aforementioned competition were chosen to test the strengths of U-scores in a competitive environment. The test suite included easy functions for which we expected that trials would terminate at EVmin, functions of moderate difficulty for which trials could end on both terminal axes and very difficult functions for which successes, if any, should be rare. All functions were shifted to thwart algorithms that preferentially search for solutions at the origin and five functions were fully rotated to ensure that they were fully parameter-dependent. Specifically, the functions in the benchmark suite were:

- Unimodal Function
    1. Shifted and Fully Rotated Zakharov Function.
- Basic Functions
    1. Shifted and Fully Rotated Rosenbrock's Function.
    2. Shifted and Fully Rotated Expanded Schaffer's f6 Function.
    3. Shifted and Fully Rotated Non-Continuous Rastrigin Function.
    4. Shifted and Fully Rotated Levy Function
- Hybrid Functions
    1. Hybrid Function 1 (N = 3)
        a) Bent Cigar Function
        b) HGBat Function
        c) Rastrigin's Function
    2. Hybrid Function 2 (N = 6)
        a) HGBat Function
        b) Katsuura Function
        c) Ackley's Function
        d) Rastrigin's Function
        e) Modified Schwefel's Function
        f) Schaffer's F7 Function
    3. Hybrid Function 3 (N = 5)
        a) Katsuura Function
        b) HappyCat Function
        c) Expanded Griewank's plus Rosenbrock's Function
        d) Modified Schwefel's Function
        e) Ackley's Function
- Composition Functions
    1. Composition Function 1 (N = 3)
        a) Rotated Rosenbrock's Function
        b) High Conditioned Elliptic Function
        c) Rotated Bent Cigar Function
        d) Rotated Discuss Function
        e) High Conditioned Elliptic Function
    2. Composition Function 2 (N = 3)
        a) Rotated Schwefel's Function
        b) Rotated Rastrigin's Function
        c) HGBat Function
    3. Composition Function 3 (N = 5)
        a) Expanded Schaffer's Function
        b) Modified Schwefel's Function
        c) Rosenbrock's Function

d) Rastrigin's Function
       e) Griewank's Function
    4. Composition Function 4 (N = 6)
       a) HGBat Function
       b) Rastrigin's Function
       c) Modified Schwefel's Function
       d) Bent Cigar Function
       e) High Conditioned Elliptic Function
       f) Expanded Schaffer's Function

## 5.2 Participating Algorithms

The algorithms that participated in the CEC 2022 competition are listed below according to their rank (best = 1). We assigned an abbreviated name to each algorithm to improve the readability of Table 7.

1) EA4: Eigen Crossover in Cooperative Model of Evolutionary Algorithms Applied to CEC 2022 Single Objective Numerical Optimization [27].
2) NL-LBC: NL-SHADE-LBC algorithm with linear parameter adaptation bias change for CEC 2022 Numerical Optimization [28].
3) NL-MID: A Version of NL-SHADE-RSP Algorithm with Midpoint for CEC 2022 Single Objective Bound Constrained Problems [29].
4) S-DP: Dynamic Perturbation for Population Diversity Management in Differential Evolution [30].
5) jSObin: An adaptive variant of jSO with multiple crossover strategies employing Eigen transformation [31].
6) MTT: Multiple Topology SHADE with Tolerance-based Composite Framework for CEC2022 Single Objective Bound Constrained Numerical Optimization [32].
7) IUMO: an Improved IMODE algorithm based on the Reinforcement Learning [33].
8) IMPML: Improvement-of-Multi-Population ML-SHADE [34].
9) NLSOMA: NL-SOMA-CLP for Real Parameter Single Objective Bound Constrained Optimization [35].
10) ZOCMAES: Zeroth-Order Covariance Matrix Adaptation Evolution Strategy for Single Objective Bound Constrained Numerical Optimization Competition [36].
11) OMCSO: Opposite Learning and Multi-Migrating Strategy-Based Self-Organizing Migrating Algorithm with the Convergence Monitoring Mechanism [37].
12) Co-PPSO: Performance of Composite PPSO on Single Objective Bound Constrained Numerical Optimization Problems of CEC 2022 [38].
13) SPHH: An ensemble of single point selection perturbative hyper-heuristics [39].

## 5.3 Competition Results

Table 7 presents the results of the 2022 CEC Competition on Single Objective Bound-Constrained Numerical Optimization. An algorithm's final score ("Total" in Table 7) was the *unweighted* sum of its twelve 10-dimensional and twelve 20-dimensional U-scores. Figure 8 gives a more illustrative view of the results.

**Table 7.** Algorithms' ranks and their U-scores. Best is in **bold**.

| Rank | Algorithm | 10-$D$ | 20-$D$ | Total |
|---|---|---|---|---|
| 1 | EA4 | 92239 | **94139.5** | **186378.5** |
| 2 | NL-LBC | **95230.5** | 86685 | 181915.5 |
| 3 | NL-MID | 80283 | 75188.5 | 155471.5 |
| 4 | S-DP | 65823.5 | 78219.5 | 144043 |
| 5 | jSObin | 65348 | 77421 | 142769 |
| 6 | MTT | 71860.5 | 69258.5 | 141119 |
| 7 | IUMO | 77085.5 | 62444 | 139529.5 |
| 8 | IMPML | 57460 | 65457 | 122917 |
| 9 | NLSOMA | 53927 | 59696.5 | 113623.5 |
| 10 | ZOCMAES | 50027 | 57687 | 107714 |
| 11 | OMCSO | 48571.5 | 51877 | 100448.5 |
| 12 | Co-PPSO | 38092.5 | 34195.5 | 72288 |
| 13 | SPHH | 40337 | 14738 | 55075 |

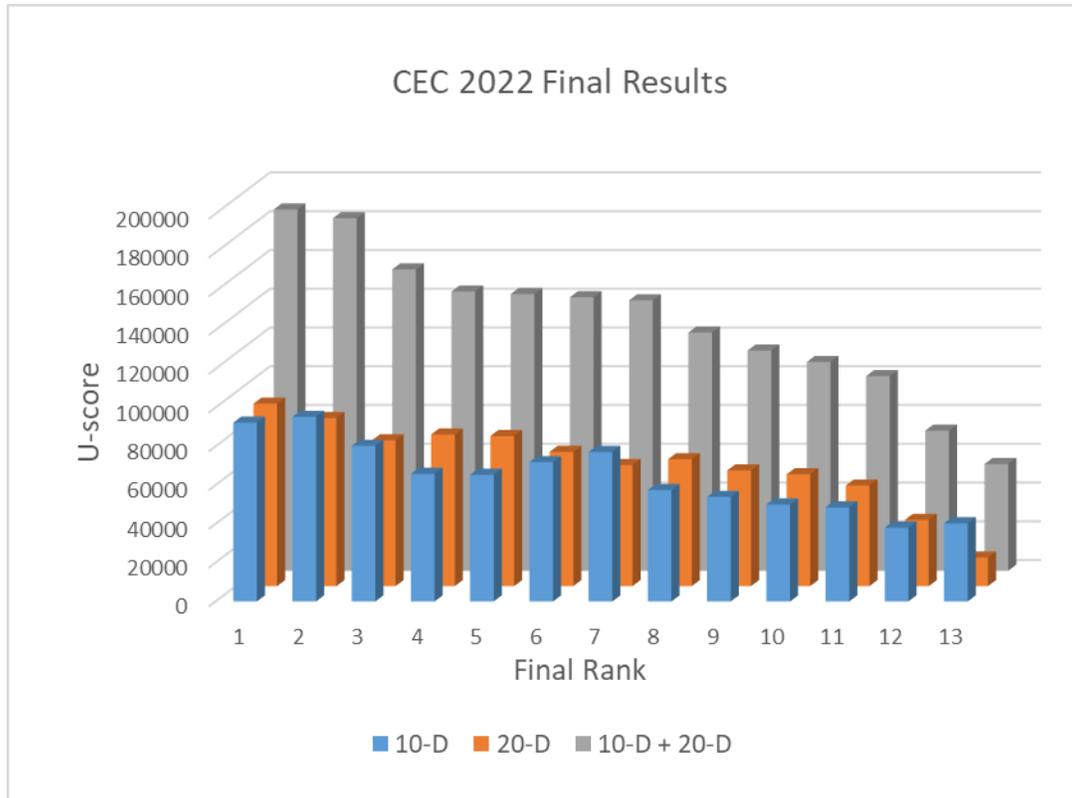

**Fig. 8**. U-scores for the 10-*D* (front), 20-*D* (middle) and combined 10-*D* + 20-*D* results (back) for the 2022 CEC Special Session and Competition on Single Objective Bound Constrained Numerical Optimization.

While there were no ties among algorithms, the appearance of 0.5 in some of the scores indicates that there were ties among some trials. These occurred at FEmax after trials had reached the same non-optimal solution (to within a difference in error value of 1.0e−8). To distinguish between these sub-optimal trials, it would be necessary to consider U-scores sampled at multiple FE cut-points— not just at FEmax—so that the algorithm that first reached the non-optimal basin could be rewarded.

Because there are $n^2$ comparisons between each pair of algorithms and $m(m - 1)/2$ pairs of algorithms, the maximum U-score for each function is $n^2 m(m - 1)/2$. Consequently normalization is not necessary to equalize each function's contribution to the final score. This makes it is easy to compute the effect that a dimension-based function weight *would* have had on the final rankings. As shown in Table 8, a weight of 1.0 for both the 10- and 20-dimensional functions reproduces the original rankings, whereas multiplying the 20-dimensional U-scores by a weight of 4.77 causes the third and fourth ranked algorithms to swap places. The algorithms ranked seventh and eighth swap places once the weight reaches 6.51 and a slightly higher weight of 6.69 swaps the fifth and third place algorithms. Any function weight greater than 6.69 and the rankings only depend on the results for the 20-dimensional functions, i.e. the 10-dimensional results do not affect the rankings.

**Table 8.** Hypothetical U-score weights for 20-dimensional functions and their effect on ranks. The weight for 10-*D* functions is 1.0.

| Weight | Rank | | | | | | | | | | | | |
|---|---|---|---|---|---|---|---|---|---|---|---|---|---|
| 1.0 | 1 | 2 | 3 | 4 | 5 | 6 | 7 | 8 | 9 | 10 | 11 | 12 | 13 |
| 4.77 | 1 | 2 | 4 | 3 | 5 | 6 | 7 | 8 | 9 | 10 | 11 | 12 | 13 |
| 6.51 | 1 | 2 | 4 | 3 | 5 | 6 | 8 | 7 | 9 | 10 | 11 | 12 | 13 |
| $\geq 6.69$ | 1 | 2 | 4 | 5 | 3 | 6 | 8 | 7 | 9 | 10 | 11 | 12 | 13 |

The difference in the combined 10- and 20-dimensional U-scores of the fourth, fifth, sixth and seventh algorithms was very small. While clearly distinguished by their U-scores, it seems unlikely that dominance based on average speed and average error would have been able to do so. It does, however, also seem likely that the difference in their U-scores is not statistically significant.

## 6. Conclusion

The performance of stochastic optimization algorithms is characterized by both the speed and accuracy of multiple trials. Transforming this raw data into a metric that can determine which of two algorithms performs better on a given function is problematic because speed and accuracy are conflicting objectives. In particular, dominance that is based on the relationship between averages of speed and accuracy fails to distinguish the better algorithm in a significant portion of cases.

Instead of relying on trial averages, we can determine the better algorithm by exploiting the dominance relation that exists between individual trials. Once trials have been ranked by dominance, the better of two algorithms will be the one with the higher Mann-Whitney U-statistic, i.e. the higher number of wins when all trials are compared.

There are multiple advantages to U-scores. Perhaps most importantly, U-scores overcome the limitations inherent in the individual fixed-cost and fixed-target scenarios to reward both an algorithm's speed and its accuracy. Furthermore, U-scores account for of *all* of an algorithm's trials. Additionally:

- U-scores are based on the Mann-Whitney U-test, which is already a proven non-parametric method for comparing algorithmic performance when there is a single variable.
- Unlike average-based dominance, increasing the number of trials not only increases the granularity of the dominance classes, but also reduces the already low probability of a tie.
- U-scores merge an algorithm's comparative speed and accuracy into a single number *without making any assumptions* about their relative importance.

- Even though they do not measure either speed or accuracy directly, U-scores do give a sense of *how much* better one algorithm is than another in the sense that they represent the number of wins when all trials are compared.
- U-scores automatically assign all functions the same maximum score.
- With the same dataset, U-scores can also rank functions based on how difficult they were for the given algorithms.
- Both simulations and data from the CEC 2022 competition have demonstrated that U-scores are effective.

This paper's key insight is that trials from multiple algorithms can be linearly ordered regardless of how they terminate. Consequently, any test statistic that can be applied to ordinal data becomes a potential score. In this paper, we based algorithmic function scores on the Mann-Whitney U-statistic, primarily because it has a straightforward interpretation as the number of wins when all trials are compared. We could have chosen the Wilcoxon rank-sum statistic, but because its sum-of-ranks statistic is related to the number of wins by a simple constant, it would not have changed the algorithms' rankings. Alternatively, scores based on the Friedman test's Q statistic for example, may rank algorithms differently than U-scores. There are many possibilities.

Like all measures of performance, the U-score makes compromises, e.g. it is a relative measure, but experiments with both real and simulated data show that it is a fair and effective and way to compare the speed *and* accuracy of stochastic optimization algorithms.